# EŞLE/İNDİRGE YÖNTEMİ KULLANILARAK DESTEK VEKTÖR MAKİNESİ ALGORİTMASI İLE YÜKSEK BOYUTLU SOSYAL MEDYA MESAJLARININ KUTUPSAL DEĞERİNİN ÖLÇÜLMESİ

Ferhat Özgür Çatak
TÜBİTAK – BİLGEM – SİBER GÜVENLİK ENSTİTÜSÜ
ozgur.catak@tubitak.gov.tr

## ÖZET

Bu çalışmada önerilen yöntem kullanılarak, Eşle/İndirge (MapReduce) tekniği ile özellikle TFxIDF yöntemi gibi yüksek boyutlu veri setlerinin, veri madenciliğinde oldukça sık kullanılan makine öğrenme algoritmalarından olan Destek Vektör Makinesi (DVM) ile uygulanabilirliğini anlatılmaktadır. Literatürde, DVM sınıflandırma algoritması, makine öğrenmesi yöntemleri arasında genelleştirme özelliği en yüksek ve nitelikler arasında yer alan korelasyondan etkilenmeyen bir yöntem olduğunu gösteren birçok çalışma mevcuttur. Fakat DVM yöntemi eğitim aşamasında kuadratik optimizasyon yöntemleri kullanmasından dolayı zaman karmaşıklığı $O(m^3)$ ve alan karmaşıklığı $O(m^2)$ şeklindedir. Bu nedenle DVM, yüksek boyutlu veri setlerinin sınıflandırılmasında kullanılacak hipotezin çıkarımı esnasında uygulanabilir olmaktan çıkmaktadır. Bu soruna çözüm olarak geliştirilen yöntemde, dağıtık Eşle/İndirge yöntemiyle alt veri setlerinin oluşturulması, her bir alt veri seti kullanılarak ortaya çıkan sınıflandırma hipotezinin destek vektörlerinin birleşimi, yinelemeli olarak tekrar kullanımıyla eğitilmesi anlatılmaktadır. Çalışmanın uygulama kısmında, yüksek boyutlu sosyal medya mesaj veri setinin TFxIDF yöntemi ile gösterimi ve bu sayısal veri setinin duygu analizi (sentiment analysis) ile kutupsal değerinin ölçümü anlatılmaktadır. Sınıflandırma yöntemi olarak iki ve üç sınıflı modeller oluşturulmuştur. Her bir sınıflandırma modelinin karmaşıklık matrisi tablolar ile gösterilmiştir Sosyal medya mesaj derlemi Türkiye'de bulunan 108 devlet ve 66 adet vakıf üniversitesi mesajlardan oluşmaktadır. Derlem için kaynak olarak Twitter kullanılmıştır. Twitter kullanıcıların mesajları Twitter Streaming API ile elde edilmiştir. Sonuçlar grafik ve tablolar ile paylaşılmıştır.
**Anahtar Kelimeler:** Büyük Veri, Destek Vektör Makinesi, Eşle/İndirge, Makine Öğrenmesi, Sosyal Medya

# POLARIZATION MEASUREMENT OF HIGH DIMENSIONAL SOCIAL MEDIA MESSAGES WITH SUPPORT VECTOR MACHINE ALGORITHM USING MAPREDUCE

## SUMMARY

In this article, we propose a new Support Vector Machine (SVM) training algorithm based on distributed MapReduce technique. In literature, there are a lots of research that shows us SVM has highest generalization property among classification algorithms used in machine learning area. Also, SVM classifier model is not affected by correlations of the features. But SVM uses quadratic optimization techniques in its training phase. The SVM algorithm is formulated as quadratic optimization problem. Quadratic optimization problem has $O(m^3)$ time and $O(m^2)$ space complexity, where $m$ is the training set size. The computation time of SVM training is quadratic in the number of training instances. In this reason, SVM is not a suitable classification algorithm for large scale dataset classification. To solve this training problem we developed a new distributed MapReduce method developed. Accordingly, (i) SVM algorithm is trained in distributed dataset individually; (ii) then merge all support vectors of classifier model in every trained node; and (iii) iterate these two steps until the classifier model converges to the optimal classifier function. In the implementation phase, large scale social media dataset is presented in TFxIDF matrix. The matrix is used for sentiment analysis to get polarization value. Two and three class models are created for classification method. Confusion matrices of each classification model are presented in tables. Social media messages corpus consists of 108 public and 66 private universities messages in Turkey. Twitter is used for source of corpus. Twitter user messages are collected using Twitter Streaming API. Results are shown in graphics and tables.

## GİRİŞ

Sosyal medyanın kullanımının gelişmesi ile beraber bilgiye erişim oldukça kolaylaşmıştır. Sosyal medyanın karşılıklı iletişime izin veren yapısından dolayı popülerliği artmaktadır (Thelwall, 2008).

Duygu analizi (Sentiment Analysis), bir metin parçasının herhangi bir konu hakkında duygu veya düşünce içerdiğini ve içermesi durumunda bu metinin kutupsal değerini ölçmek için kullanılan otomatik bir süreçtir (Paltoglou & Thelwall, 2012). Duygu analizi yönteminin günümüzde oldukça sık kullanılmasının en önemli nedeni, internet ortamında kullanıcılar tarafından oluşturulan içeriğin artmasıyla kurumlar ve şirketler için sağlıklı bilgilerin çıkarılabilmesi isteği olmuştur. Bu konuda yapılan ilk çalışmalar, herhangi bir ürün hakkında yapılan çalışmalar olmuştur (Littman & Turney, 2002; Pang & Lee, 2008).

Sosyal medyada yer alan metinleri üzerinde duygu analizi işlemi, metin sınıflandırma işleminin özel bir alanıdır. Metin sınıflandırma işlemleri oldukça karmaşık ve doğası gereği oldukça yüksek sayıda nitelik içermektedir (Yang & Pedersen, 1997). Türkçe dili için Çetin ve Amasyalı (Çetin & Amasyalı, 2013) yaptıkları çalışmada sosyal medya üzerinde bir Telekom şirketi için olumlu, olumsuz ve nötr şeklinde üç farklı sınıfı ayırmışlardır. Kullandıkları veri seti 6000 adet sosyal medya mesajından oluşmaktadır. Pozitif sınıfa ait 3040 adet mesaj, negatif sınıfa ait 1847 adet mesaj ve nötr sınıfa ait 1113 adet mesaj içermektedir. En iyi sınıflandırmaya % 65,5 doğruluk oranı ile DVM algoritması kullanarak erişmişlerdir. İngilizce dili için 2013 yılında Ghiassi ve diğerleri (Ghiassi, Skinner, & Zimbra, 2013) sosyal medya üzerinde yer alan mesajları kullanarak marka bilinirlikleri için bir çalışma yapmışlardır. İstatiksel analiz yöntemleri ve eğiticili nitelik çıkarım yöntemlerini kullanarak veri seti oluşturmuşlardır. Oluşan veri seti kullanılarak dinamik yapay sinir ağları ve DVM sınıflandırma algoritmaları ile sınıflandırma yaparak % 83,9 doğruluk oranına ulaşmışlardır. İngilizce dili için başka bir çalışma 2013 yılında Urena-Lopez ve diğerleri (Montejo-Ráez, Martínez-Cámara, Martín-Valdivia, & Urena-López, 2013) sosyal medya üzerinde yer alan mesajların düşüncelerin kutupsal sınıflandırması için yaptıkları çalışmada WordNet ağında yer alan düğümlerin ağırlık vektörlerini çıkarmışlardır. DVM algoritması ile eğiticili sınıflandırma yöntemi kullanılmış % 64.29 doğruluk oranına ulaşmışlardır. Kullandıkları veri seti toplam 376.296 adet sosyal medya mesajı içermektedir.

## DESTEK VEKTÖR MAKİNESİ

DVM sınıflandırma algoritması, sınıflandırma modelinin yüksek genelleştirme özelliğinden nedeniyle makine öğrenmesi alanında yaygın olarak kullanılmaktadır (Çatak & Balaban, 2013). DVM, veri setinin sahip olduğu veya daha yüksek boyutlu bir uzayda yer alan örnekleri en uygun şekilde ikiye ayıran bir hiperdüzlem oluşturmaktadır (Vapnik, 1995). Yumuşak marjin DVM sınıflandırma algoritması, marjin maksimizasyonunu uygularken hatalı sınıflandırılan örneklere izin vermektedir (Cortes & Vapnik, 1995). Bu yöntem gevşek değişkenler $\xi_i$ ile $\vec{x}_i$ örneğinin hatalı sınıflandırmayı ölçen değişkenler kullanmaktadır. Bu değişken kullanılarak yeni kısıt şu şekilde olacaktır.

$$y_i(\vec{w}.\vec{x}_i + b) \geq 1 - \xi_i, \qquad \xi_i \geq 0, \ \ i = 1, \dots, n \tag{1}$$

Yumuşak marjin optimizasyon problemi şu şekle dönüşmektedir.

$$minimize: f(\vec{w}, b, \xi_i) = \frac{1}{2}\|w\|^2 + C \sum_i \xi_i \tag{2}$$

$$kısıtlar: y_i(\vec{w}.\vec{x}_i + b) \geq 1 - \xi_i, \quad \xi_i \geq 0, \quad 0 \leq \alpha_i$$

## EŞLE/İNDİRGE (MAPREDUCE)

Yüksek boyutlu veri seti alt kümelere parçalanması ile büyük problemin daha ufak alt problemlere parçalanması hedeflenmektedir (Amdah, 1967). Fakat bu durumda ortaya yeni problemler çıkmaktadır. Genel olarak bu problemler şu şekilde özetlenebilir (Lin & Dyer, 2010).

- Büyük problem, paralel çalışabilecek küçük alt görevler içeren alt problemlere nasıl ayrıştırılabilir?
- Dağıtık yapıda bilgisayarlara kısıtlar ve kaynakları göz önünde bulundurularak bu görevlere nasıl atanabilir?
- Bu dağıtık bilgisayarlar arasında ki senkronizasyon nasıl sağlanabilir?
- Dağıtık sistem üzerinde yer alan yazılım hataları ve donanım arızaları durumunda ilgili bilgisayara atanmış olan görevler nasıl devam edebilir?

Kullanıcılar tarafından tanımlanan eşle (map) ve indirge (reduce) fonksiyonları ve bu fonksiyonlara girdi değeri olarak verilen çağrışımsal diziler kullanılmaktadır. Anahtar-değer ikilileri integer, float, string gibi veri tipleri olabileceği gibi liste, değişkenler grubu gibi karmaşık yapılarda olabilmektedir (Lin & Dyer, 2010). Eşle fonksiyonu bir veri alanında bulunan veri çiftlerini alarak bunları farklı bir alana veri çift listesi olarak vermektedir.

$$eşle(a_1, d_1) \rightarrow liste(a_2, d_2) \tag{3}$$

İndirge fonksiyonu ise eşle fonksiyonu çıktı değerlerine uygulayarak yeni liste oluşturmaktadır.

$$indirge(a_1, liste(d_1)) \rightarrow liste(a_2, d_2) \tag{4}$$

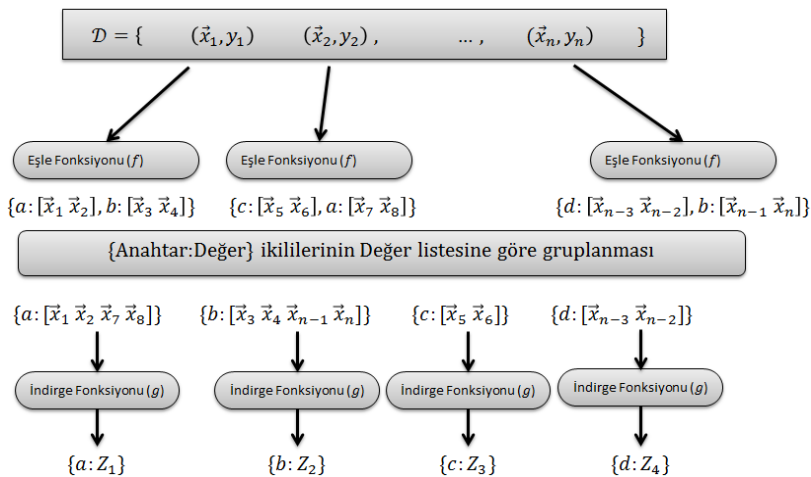

**Şekil 2:** Eşle/İndirge tekniğinin genel gösterimi.

Eşle/İndirge çatısının anahtar/değer şeklindeki çiftlerden oluşan listeyi değerler listesi şekline çevirmektedir. Şekil 2'de Eşle/İndirge tekniği genel gösterimi bulunmaktadır. İki aşamadan oluşan bu yöntem ile çıktı olarak genellikle başka bir Eşle/İndirge işleminde kullanılmak üzere kullanılacak $n$ adet matris ortaya çıkmaktadır.

## EŞLE/İNDİRGE DVM ALGORİTMASININ TASARIMI

Geliştirilen model, bütün veri setini alt veri setlerine parçalamakta ve bu alt veri setleri Eşle/İndirge yönteminin eşle fonksiyonuna atanmaktadır. Her bir eşle fonksiyonu global destek vektör matrisi ile kendisine atanmış olan alt veri seti birleştirmektedir. Bu yeni alt veri seti, eşle fonksiyonunun değeri anahtar bilgisi olacak şekilde indirge fonksiyonuna parametre olarak verilmektedir. Paralel bir yapıda çalışmış olan indirge fonksiyonu kendisine girdi olarak verilen alt veri setini DVM ile sınıflandırıcı hipotez bulmaktadır. Paralel çalışmış olan indirge fonksiyonu çıktı olarak $\alpha > 0$ olan alt veri seti örneklerini $S_l$, destek vektörlerini ve sınıflandırıcı hipotezi vermektedir. Bu sistemin genel yapısı Şekil 3'de gösterilmektedir.

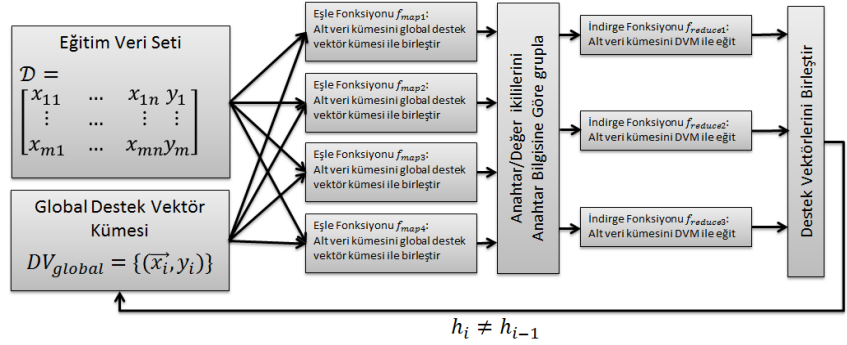

**Şekil 3:** Eşle/İndirge DVM algoritmasının genel yapısı.

Eşle/İndirge işleminin ilk aşaması olan eşle fonksiyonunda, eğitim veri setinin parçalanmış olan alt kümesi global destek vektör kümesi ile birleştirilmektedir. İşlemin ikinci fonksiyonu olan indirge aşamasında ise bu yeni alt veri seti kümesi DVM sınıflandırma algoritması ile eğitilmektedir. Tüm İndirge fonksiyonlarından çıkan destek vektörleri birleştirilerek oluşturulan matrisi ile global destek vektör matrisi birleştirilerek tekrar global destek vektör matrisi olarak saklanmaktadır. Eşle ve İndirge fonksiyonlarında kullanılan algoritmanın sözdekodu (Pseudcode) Tablo 1 ve Tablo 2'de gösterilmektedir.

**Tablo 1:** İkili sınıf destek vektör makinesi Eşle fonksiyonu.

---
**Algoritma 1** İkili sınıf destek vektör makinesi Eşle fonksiyonu

---

$\overrightarrow{DV}_{Global} = \emptyset$ // *Global destek vektör matrisini temizle*
**while** $h^t \neq h^{t-1}$
  **for** $l \in L$ **do** // *her bir alt veri seti için*
    $D_l^t \leftarrow D_l^t \cup DV_{Global}^t$
  **end for**
**end while**

---

**Tablo 2:** İkili sınıf destek vektör makinesi İndirge fonksiyonu.

---
**Algoritma 2** İkili sınıf destek vektör makinesi İndirge fonksiyonu

---

**while** $h^t \neq h^{t-1}$ **do**
  **for** $l \in L$
// *birleştirilmiş veri setini eğit*

```
// destek vektörlerini ve ikili-sınıf hipotezi bul
    DV_l, h^t ← binarySvm(D_l)
  end for
  for l ∈ L
    DV_{Global} ← DV_{Global} ∪ SV_l
  end for
end while
```

$X$ veri seti kümesi olsun. Veri seti şu formda gösterilebilir.

$$X = \left\{ (\vec{x}_i, y) \mid \vec{x}_i \in \mathbb{R}^{l \times n}, y \in \{-1, +1\} \right\}_{i=1}^{n}$$

$X$ veri seti kümesinin alt kümesi $S$ olsun. $F(S)$ ise $S$ alt veri seti için optimal hedef fonksiyonumuz, $h^*$ ise mimimum deneysel riske ($R_{Deneysel}(h)$) sahip olan global optimal hipotezi göstermektedir. Çalışmada geliştirilen modelde, yineleme $t = 0$ iken global destek vektörü $SV_g^0 = \emptyset$'dir. Global destek vektörü, $SV_g^t$, $t$ yinelemesinde bulunan her bir indirge fonksiyonunda ortaya çıkan destek vektörü kümesinin birleşimini göstermektedir. Kayıp fonksiyonu kullanılarak sınıflandırıcı modelin deneysel risk değeri şu şekilde hesaplanmaktadır.

$$R_{Deneysel}(h) = \frac{1}{n} \sum_{i=1}^{n} \ell(h(x), y) \tag{6}$$

Deneysel riskin minimize edilmesi prensibine göre, öğrenme algoritması, $\mathcal{H}$ hipotez uzayında yer alan sınıflandırıcı modeller arasında deneysel riski en düşük olan $h^*$ sınıflandırıcı modelini seçmelidir.

$$h^* = \arg \min_{h \in \mathcal{H}} R_{Deneysel}(h) \tag{7}$$

Çalışmada geliştirilen modelde $t$ iterasyonunda indirge fonksiyonu sırasında bulduğu hipotez uzayında, deneysel riski en düşük olan sınıflandırıcı $h^t$ şeklinde gösterilmektedir. Model, en düşük deneysel riskin değişimine göre kendini sonlandırmaktadır. Algoritmanın durması için gerekli olan koşul şu şekildedir.

$$\left| R_{Deneysel}(h^{t-1}) - R_{Deneysel}(h^t) \right| \leq \gamma, \ \gamma \geq 0 \tag{8}$$

Her bir yineleme sonunda, alt parçalara ayrılmış olan veri seti, global destek vektör seti ile birleştirilerek güçlendirilmektedir. İstatiksel öğrenme teorisinde de kullanılan büyük sayılar kuralına deneysel risk değeri veri seti güçlendirildikçe beklenen değere yaklaşacaktır.

$$S \hookleftarrow S \cup (SV_{Global}^t) \tag{9}$$

Alt veri seti $S$ kümesine global destek vektörlerini ekleyerek güçlendirilerek sınıflandırıcı fonksiyonun doğruluk oranı artmaktadır.

# UYGULAMA SÜRECİ

Çalışmada Türkiye'de bulunan 108 adet devlet üniversitesi ve 66 adet vakıf üniversitesi olmak üzere ele alınmıştır. Sınıflandırma modelinin oluşturulması için üniversiteler sosyal medya mesaj veri setinden farklı olarak 3.404.074 mesaj kullanılmıştır. Uygulama için Java tabanlı bir uygulama geliştirilerek Twitter Stream API v1.1 kullanılmıştır. Üniversite adlarından ve resmi Twitter hesaplarından oluşan bir kelime vektörü oluşturulmuştur. Oluşturulan bu kelime vektörü ile Twitter4j API bileşeni kullanılarak Twitter izlenmeye başlanmıştır. Mesajlar TFxIDF matrisleri oluşturularak sayısal hale dönüştürülmüştür. Sayısal veri seti Eşle/İndirge tabanlı DVM sınıflandırma algoritmasında eğitilmiş ortaya çıkan model sayısallaştırılmış veri setlerinin sınıflandırılmasında kullanılmıştır.

**Veri Seti Üzerinde Yapılan İşlemler**

Oluşturulan bu veri seti sırasıyla aşağıdaki aşamalardan geçmiştir:

- Bir dilde çok kullanılan ve genellikle doğal dilde göz ardı edilen kelimelere etkisiz kelimeler denilmektedir. Örnek olarak "ama", "bazı", "nasıl" verilebilir. Bu kelimelerin listesi Tablo 4'de gösterilmiştir. Bu kelimeler veri setinden temizlenmiştir.
- Metin şeklinde olan veri seti vektör uzayına taşınmıştır.
- Oluşan vektör uzayından nitelik seçimi yapılmıştır.
- Oluşan veri seti DVM sınıflandırma algoritması ile eğitilmiştir. Ortaya çıkan sınıflandırma modeli üniversite mesajlarının ayrıştırılmasında kullanılmıştır.

**Tablo 4:** Çalışmada kullanılan ve Türkçe' de cümlelere anlam katmayan kelimelerin listesi.

| acaba, altı, altmış, ama, bana, bazı, belki, ben, benden, beni, benim, beş, bi, bin, bir, biri, birkaç, birkez, birşey, birşeyi, biz, bizden, bizi, bizim, bu, buna, bunda, bundan, bunu, bunun, çok, çünkü, çünkü, da, daha, dahi, de, defa, diye, doksan, dokuz, dört, elli, en, gibi, hem, hep, hepsi, her, hiç, için, iki, ile, ise, katrilyon, kez, kırk, ki, kim, kimden, kime, kimi, mı, milyar, milyon, mu, mü, nasıl, ne, neden, nerde, nerede, nereye, niçin, niye, on, ona, ondan, onlar, onlardan, onların, onlari, onu, otuz, sanki, sekiz, seksen, sen, senden, seni, senin, siz, sizden, sizi, sizin, şey, şeyden, şeyi, şeyler, şu, şuna, şunda, şundan, şunu, trilyon, tüm, üç, ve, veya, ya, yani, yedi, yetmiş, yirmi, yüz |
|---|

**Veri Setinin Vektör Uzayına Dönüştürülmesi**

Veri setinin vektör uzayına çıkarılması işleminde birinci adım mesajlardan niteliklerin çıkarılmasıdır. Bir doküman $d$ içerisinde yer alan her bir terim $t$'ye, bu dokümanda görülme sıklığına göre bir ağırlık atama işlemi yapılır (Manning, Raghavan, & Schütze, 2008). Bu işleme terim frekansı, $tf_{t,d}$ denilmektedir. Ters doküman frekansı derlem içerisinde bulunan doküman sayısının, $N$, doküman frekansına, $df_t$, bölümünün logaritması olarak tanımlanmaktadır.

$$idf_t = log \frac{N}{df_t} \tag{10}$$

Her bir dokümanda yer alan terimlerin ağırlıklarını hesaplamak için terim frekansı $tf_{t,d}$ ve ters doküman frekansı $idf_t$ birleştirilerek terim frekansı-ters doküman frekansı (TF-IDF) matrisi oluşturulur.

$$tf\ idf_t = tf_{t,d} \times idf_t \tag{11}$$

**Veri Setinin Hazırlanması**

Bu çalışma kapsamında, DVM sınıflandırma algoritması eğitim aşamasında kullanılacak iki sınıflı sınıflandırma modelinin çıkarılması için 347.158 adet sosyal medya mesajı, üç sınıflı sınıflandırma modelinin çıkarılması için 335.070 adet sosyal medya mesajı seçilmiştir. Yapılan uygulamada, DVM sınıflandırma algoritmasının eğitiminde kullanılan veri setinin iki ve üç sınıf için mesaj sayısı Tablo 5'de gösterilmektedir.

**Tablo 5:** Sınıflandırma veri seti mesaj dağılımı.

|  | **Olumlu** | **Olumsuz** |  | **Olumlu** | **Olumsuz** | **Nötr** |
|---|---|---|---|---|---|---|
| **İkili Sınıf** | 174.669 | 172.489 | **Üç** | 113.438 | 111.779 | 109.853 |

Eğitim veri setleri ile oluşturulan DVM sınıflandırma modelleri üniversiteler veri setine sınıflandırma için uygulanmıştır.

**İki Sınıflı Sınıflandırma Modeli Sonuçları**

Üniversiteler mesaj veri seti "Olumlu", "Olumsuz" şeklinde sınıflandırılmıştır. Oluşturulan DVM sınıflandırma modelinin karmaşıklık matrisi Tablo 6'da gösterilmiştir.

**Tablo 6:** İki sınıflı DVM sınıflandırma modelinin karmaşıklık matrisi.

|  |  | **Tahmin Sınıf** | |
|---|---|---|---|
|  |  | -1 | 1 |
| **Gerçek Sınıf** | -1 | % 40.61 | % 9.03 |
|  | 1 | % 5.04 | % 45.31 |

Elde edilen sonuçlara göre en fazla mesaja sahip olan ilk 10 üniversite, olumlu mesaj oranı ve olumsuz mesaj oranına göre yapılan sıralamalar ile ikili sınıflandırmanın sonuçları Tablo 7'de gösterilmiştir.

**Tablo 7:** Mesaj sayısına, olumlu ve olumsuz mesaj sayısına göre ilk 10 üniversitenin iki sınıflı sınıflandırma sonuçları.

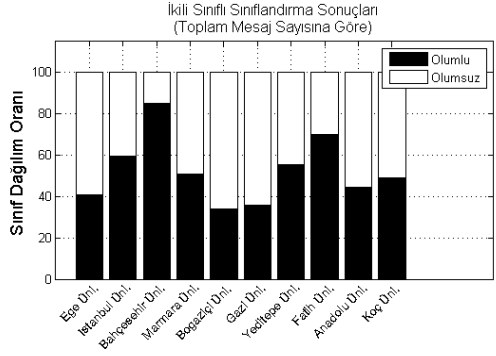
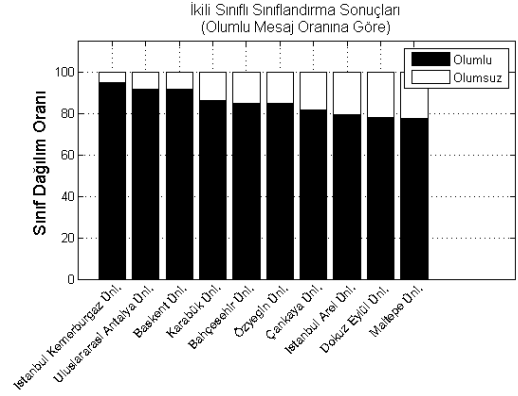
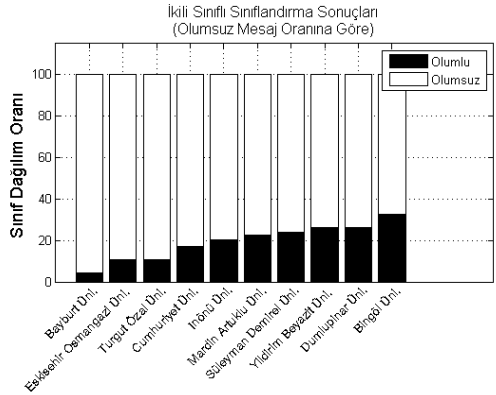

## Üç Sınıflı Sınıflandırma Modeli Sonuçları

Üniversiteler mesaj veri seti "Olumlu", "Olumsuz" ve "Nötr" şeklinde sınıflandırılmıştır. Oluşturulan DVM sınıflandırma modelin karmaşıklık matrisi Tablo 8'de gösterilmiştir.

**Tablo 8:** Üç sınıflı DVM sınıflandırma modelinin karmaşıklık matrisi.

|  |  | **Tahmin Sınıf** | | |
|---|---|---|---|---|
|  |  | -1 | 0 | 1 |
| **Gerçek Sınıf** | -1 | % 23,63 | % 6,24 | % 3,25 |
|  | 0 | % 3,44 | % 21,47 | % 8,06 |
|  | 1 | % 2,16 | % 8,46 | % 23,28 |

Elde edilen sonuçlara göre en fazla mesaja sahip olan ilk 10 üniversite, olumlu mesaj oranı, nötr mesaj oranı ve olumsuz mesaj oranına göre yapılan sıralamalar ile üçlü sınıflandırmanın sonuçları Tablo 9'da gösterilmiştir.

**Tablo 9:** İlk 10 üniversitenin üç sınıflı sınıflandırma sonuçları.

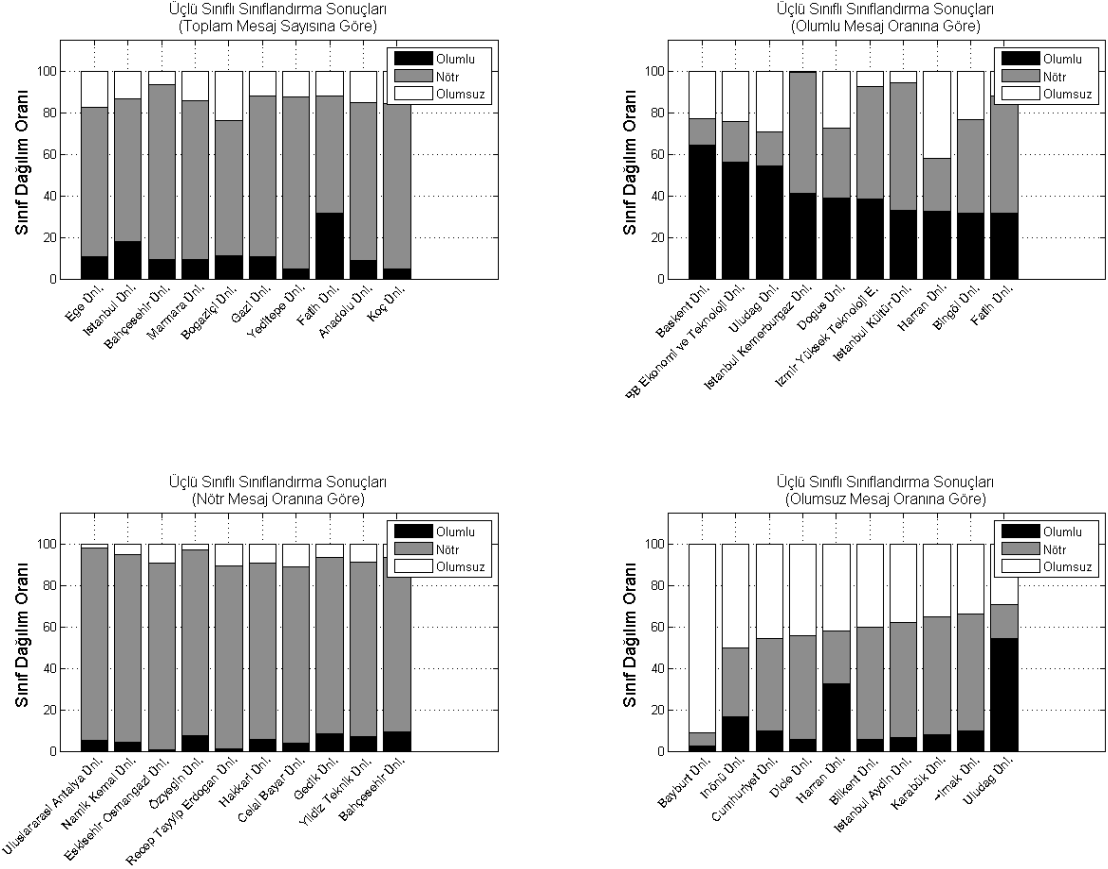

# SONUÇ

Makine öğrenmesinde kullanılan veri setleri her geçen gün artmaktadır. Büyük veri (Big Data) şeklinde tanımlanan kavram üzerinde makine öğrenme yöntemleri uygulanabilmesi için çeşitli araştırmalar yapılmaktadır (Low, ve diğerleri, 2012; Lohr, 2012; Yu, Hsieh, Chang, & Lin, 2012).

Olasılık teorisinde yer alan büyük sayılar kanununa göre, bir deney oldukça yüksek sayıda uygulandıktan sonra elde edilen değerlerin ortalaması beklenen değere oldukça yakın olması beklenir. Deneysel risk, sınıflandırma fonksiyonunu elde etmek için eğitim sırasında kullanılan veri setinin ortalama kayıp değerini, gerçek risk ise yeni örneklerde dahil olmak üzere bütün dağılımın ortalama kayıp değerini göstermektedir. Hoeffding 1963 yılında deneysel riskin beklenen değeri nasıl karakterize edeceğini, sınıflandırma fonksiyonunun $f$'nin deneysel riski, veri seti örneklem sayısı sonsuza gittikçe gerçek riske yakınsayacağı ile ifade etmiştir. Bu durumda makine öğrenmesinde yer alan bir veri setinin örneklem sayısının olabildiğince yüksek olması, sınıflandırıcı fonksiyonun riskini düşürecektir. Bu çalışmada

DVM sınıflandırma algoritmasının eğitiminde kullanacağı eğitim veri setinin dağıtık bir yapı ile artırılması hedeflenmiştir.

Bu çalışmada, ilk defa Türkiye'de bulunan vakıf ve devlet üniversiteleri ile ilgili olarak sosyal medya sitesi Twitter üzerinde gönderilen mesajları içeren veri seti hazırlanmıştır. Veri seti hazırlanırken varsayım olarak, üniversitelerin mesajlarının Twitter tarafından geliştiricilere sunulan Streaming API v1.1 ile sunduğu mesajlardır. Bu mesajlar üzerinden DVM sınıflandırma algoritması kullanılarak sınıflandırma işlemi yapılmıştır. İki ve üç sınıflı eğitim veri setleri kullanılarak yapılan sınıflandırma işlemlerinin doğruluk oranları tablolar ile verilmiştir. Hazırlanan veri seti, Türkçe dili için literatürde bulunan en yüksek boyutlu veri setidir.

Bundan sonraki çalışmalarda, oluşturulan sınıflandırma modelinin zaman içerisinde güncellenmesi gerekli olduğu düşünülmektedir. Sosyal medya üzerinde yer alan mesajların içeriği zaman içerisinde değişeceğinden dolayı zaman içerisinde kendini güncelleyen eğitim veri seti kullanılarak sınıflandırma modelinin güncelliğini koruması gereklidir.

## KAYNAKÇA